# Private Posterior distributions from Variational approximations

**Vishesh Karwa and Dan Kifer and Aleksandra B. Slavković**

**Abstract:** Privacy preserving mechanisms such as differential privacy inject additional randomness in the form of noise in the data, beyond the sampling mechanism. Ignoring this additional noise can lead to inaccurate and invalid inferences. In this paper, we incorporate the privacy mechanism explicitly into the likelihood function by treating the original data as missing, with an end goal of estimating posterior distributions over model parameters. This leads to a principled way of performing valid statistical inference using private data, however, the corresponding likelihoods are intractable. In this paper, we derive fast and accurate variational approximations to tackle such intractable likelihoods that arise due to privacy. We focus on estimating posterior distributions of parameters of the naive Bayes log-linear model, where the sufficient statistics of this model are shared using a differentially private interface. Using a simulation study, we show that the posterior approximations outperform the naive method of ignoring the noise addition mechanism.

## 1. Introduction and Summary

Privacy is a growing issue due to the availability of large scale data and it is widely accepted that to provide any meaningful privacy protection, the data sharing mechanism must introduce additional randomness into the data. Differential privacy Dwork et al. (2006a) has become one of the most popular frameworks to design such mechanisms. However, the practical use of differential privacy for performing inference in high dimensional contingency tables remains a challenge. For example, Fienberg, Rinaldo and Yang (2010) demonstrate that differentially private releases of summary statistics of high dimensional contingency tables are inconsistent with each other - there does not exist an integer valued contingency table corresponding to the released statistics and hence the noisy counts cannot be used for parameter inference. A part of this problem may be due to the fact that the privacy mechanism is usually ignored when performing inference.

More generally, let $d$ be a dataset that requires protection, and let $P(D; \theta)$ be a model on the data $D$. The end user of the private data is interested in performing inference on the parameters $\theta$. Privacy preserving mechanisms can be modeled as a family of conditional probability distributions, $P(Z|D = d, \gamma)$, i.e., the released dataset $z$ is a sample from $P(Z|D = d, \gamma)$, where the parameters of privacy mechanism $\gamma$ are known. Most of the current work advocates using a *naive* likelihood based on $P(z; \theta)$ to make inferences (either Bayesian or frequentist) about $\theta$, ignoring the privacy mechanism, with a few notable exceptions discussed in related work below. In some cases, $z$ is post-processed to minimize some form of distance from $d$, before being plugged into the *naive* likelihood, for example, see Barak et al. (2007), Hay et al. (2009). However, it has been shown that this strategy of using $z$ directly with the naive likelihood can lead to invalid and inaccurate inferences, and in many cases, the maximum likelihood and other parameter estimates may not even exist; see for example Fienberg, Rinaldo and Yang (2010), Karwa and Slavković (2012).

In this paper, we declare the original data $d$ as missing or noisy, and develop methods that incorporate the privacy mechanism into the likelihood. This ensures that the parameter estimates exist, and the statistical inference is valid. It also offers improved accuracy in estimation of $\theta$ (and $d$, if needed), and can provides meaningful estimates of standard errors. Thus one should ideally work with the likelihood $P(Z; \theta; \gamma) = \sum_d P(Z|D, \gamma) P(D; \theta)$, which requires summing over all possible missing data. In most cases, this likelihood is intractable and we need to resort to approximation methods. We use variational approximations (Jaakkola and Jordan, 2000) for performing inference in contingency tables released by a differentially private mechanism. We focus on estimating approximate posterior distributions of models when the sufficient statistics are given by two-way marginal summaries of a contingency table. The likelihood contains non-conjugate and non-differentiable terms that are not amenable to existing mean field approximations. Moreover, the parameters are constrained to lie in a simplex. We derive a new lower bound for the likelihood and use an MM algorithm Hunter and Lange (2004) to maximize the lower bound while respecting the parameter constraints. We use simulation studies to show that the new estimator based on approximate posterior distribution is more efficient than a "naive" estimator that ignores the privacy mechanism.

**Related Work:** The problem of inferring parameters from data released through privacy mechanisms has received little attention, with some notable exceptions. Most of the work focuses on post processing the noisy data to impose some form of structural constraints that exist in the non-private dataset. For example, Barak et al. (2007) develop a post processing technique to modify noisy marginal counts of a contingency table so that they are compatible with the existence of a real valued contingency table. However, Fienberg, Rinaldo and Yang (2010) show that these "post processed" counts fail to be useful for inferring parameters and fitting models - in particular, maximum likelihood estimates don't even exist. In a similar vein, Hay et al. (2009) develop a post processing technique to improve the accuracy in estimation of degree distributions. But Karwa and Slavković (2012) demonstrate that parameter estimation is not possible with these post processed counts due to non-existence of MLE. In order to resolve this issue, Karwa and Slavković (2012) develop an alternate post processing technique with an end goal of parameter estimation that requires projection on a marginal polytope defined by the model of interest. Karwa and Slavković (2015) show that this procedure leads to valid inferences - in particular, asymptotically consistent and normal parameter estimates can be obtained. In the context of network privacy, Karwa, Slavković and Krivitsky (2014) and Karwa, Krivitsky and Slavković (2015) estimate the parameters of exponential random graph models by using missing data methods and weighted MCMC to incorporate the privacy model into the likelihood. In a different but related line of work, Lin and Kifer (2013) develop an axiomatic utility framework and show that the statistical information in a private sample is maximized when the end user is modeled as a Bayesian decision maker. They illustrate this approach for estimating sorted histograms and show that it leads to improved accuracy. Finally, Williams and McSherry (2010) explore the use of variational approximations for modeling privacy mechanisms. Their variational approximation requires the conditional distribution of noisy answers to be of product form $\prod_i P(z_i|d_i)$ where $z_i$ is the noisy answer from a data point $d_i$. This requirement does not hold for many important cases, in particular when the data are released in the form of noisy sufficient statistics. Furthermore, they focus on improving prediction accuracy, whereas, crucially our focus is on parameter inference (which includes the case of accurate predictions). Finally, the lower bounds derived by Williams and McSherry (2010) depend on unknown parameters, whereas our goal is to estimate these parameters. We convert the general private inference problem, without making any independence assumptions on the conditional distributions of the noisy answers, into a sequence of optimization problems by using variational lower bounds that are then solved using techniques such as the MM algorithms Hunter and Lange (2004).

## 2. Differential Privacy

Formally, differential privacy mechanisms can be modeled as a family of conditional probability distributions, which define a distribution on the answers, conditional on the data; for a statistical overview of differential privacy, see Wasserman and Zhou (2010). Let $\delta(D, D')$ denote the Hamming distance between two datasets $D$ and $D'$. Differential Privacy is defined to limit disclosure related to presence or absence of any single individual as the following definition illustrates:

**Definition 1** (Differential Privacy). *Let $\epsilon > 0$ and $\mathcal{S}$ be the support of $\mathcal{P}$. A randomized mechanism (or a family of conditional probability distributions) $\mathcal{P}(.|D)$ is $\epsilon$-edge differentially private if*

$$\sup_{D,D',\delta(D,D')=1} \sup_{S \in \mathcal{S}} \log \frac{\mathcal{P}(S|D)}{\mathcal{P}(S|D')} \leq \epsilon,$$

In this definition, $\epsilon$ is the privacy parameter that, as we will see below, controls the amount of noise added to the query; small values of $\epsilon$ means more privacy protection. Typically $\epsilon$ is set to be smaller than 1. A basic mechanism to release the output of any function $f$ under differential privacy is the Laplace Mechanism (Dwork et al., 2006a) which adds Laplace noise proportional to the *global sensitivity* of $f$ as defined below. Let $\mathcal{D}$ be the set of all possible datasets and $||.||_1$ be the $L_1$ norm. The *global sensitivity* of a statistic $f : \mathcal{D} \to \mathbb{Z}^k$ is $GS(f) = \max_{d(D,D')=1} ||f(D) - f(D')||_1$.

One nice property of differential privacy is that any function of a differentially private mechanism is also differentially private (Dwork et al., 2006b; Wasserman and Zhou, 2010). We make use of this property since the Variational approximations that we derive can be regarded as a post-processing step of a differentially private mechanism.

## 3. Problem Setup

Let us assume that we observe a set $D$ of $N$ *iid* samples $d_1, \ldots, d_n$ of $D$ from a parametric model $P(D|\theta)$, where $\theta \in \Theta$ is a vector of parameters. Due to privacy constraints, we cannot directly see the data $D$ but instead get a sample from a privacy preserving mechanism, modeled as a conditional probability distribution $P(Z|D)$. The private data $z$ is a sample from $P(Z|D)P(D|\theta)$. Our goal is to perform inference on the parameters $\theta$ using the observed private sample $Z$, i.e we wish to infer a posterior probability distributions on the parameters $\theta$. However, the original sample $D$ is missing. Thus, we need to work with the intractable likelihood

$$L(Z;\theta) = \sum_D P(Z|D)P(D|\theta).$$

We resort to Variational approximations, Jaakkola and Jordan (2000) and derive a lower bound to the log marginal likelihood given by equation 1. To derive the Variational approximation, let $q(D)$ and $q(\theta)$ be variational distributions defined on the missing data $d$ and the unknown parameters $\theta$ respectively; these can be freely chosen. As a part of the variational approximation, we set $q(d,\theta) = q(d)q(\theta)$. The log marginal likelihood can be lower bounded as follows:

$$\log L(Z) = \log \int P(Z|D)P(D|\theta)P(\theta)d\theta dD \geq E_{q(D)q(\theta)}\left[\log \frac{P(Z|D)P(D|\theta)P(\theta)}{q(D)q(\theta)}\right]. \quad (1)$$

## 4. Private Naive Bayes Classification using Variational methods

In this section, we describe the naive Bayes model and apply variational inference to estimate the posterior distribution of the parameters of the naive Bayes model in a private manner. One of the goals in a classification problem is to learn a classifier based on a training dataset and predict the class of future observations. Let $X = (X_1, \ldots, X_K)$ be a random vector of $K$ random variables, also called *features*. Each $X_k$ takes values in $\{1, \ldots, J_k\}$. Let $Y$ be a random variable taking values in $\{1, \ldots, I\}$. $Y$ is also called a *class* variable. Let $D = (Y, X_1, \ldots, X_K)$. We observe $n$ *iid* copies of the random vector $D = (X, Y)$. Our goal is to estimate the conditional class probabilities, i.e., $P(Y|X)$ in a private manner. A naive Bayes classifier assumes that $P(X|Y) = \prod_{k=1}^{K} P(X_k|Y)$.

|   |   | $X_1$ |   |   |   | $X_2$ |   |   |   |   | $X_K$ |   |
|---|---|---|---|---|---|---|---|---|---|---|---|---|
|   |   | 1 | 2 |   |   | 1 | 2 |   |   |   | 1 | 2 |
| $Y$ | 1 | $n_{11}^1$ | $n_{12}^1$ | $Y$ | 1 | $n_{11}^2$ | $n_{12}^2$ | $\ldots$ | $Y$ | 1 | $n_{11}^K$ | $n_{12}^K$ |
|   | 2 | $n_{21}^1$ | $n_{22}^1$ |   | 2 | $n_{21}^2$ | $n_{22}^2$ |   |   | 2 | $n_{21}^K$ | $n_{22}^K$ |

TABLE 1
*Sufficient statistics of the Naive Bayes model.*

|   |   | $X_1$ |   |   |   | $X_2$ |   |   |   |   | $X_K$ |   |
|---|---|---|---|---|---|---|---|---|---|---|---|---|
|   |   | 1 | 2 |   |   | 1 | 2 |   |   |   | 1 | 2 |
| $Y$ | 1 | $p_{11}^1$ | $p_{12}^1$ | $Y$ | 1 | $p_{11}^2$ | $p_{12}^2$ | $\ldots$ | $Y$ | 1 | $p_{11}^K$ | $p_{12}^K$ |
|   | 2 | $p_{21}^1$ | $p_{22}^1$ |   | 2 | $p_{21}^2$ | $p_{22}^2$ |   |   | 2 | $p_{21}^K$ | $p_{22}^K$ |

TABLE 2
*An example of the parameters of the Naive Bayes model for a $2 \times 2 \times K$ table.*

The sufficient statistics of a naive Bayes classifier are given by the set of $K$ two-way contingency tables formed by cross classifying each feature $X_i$ with the class variable $Y$, see Table 1 for an example. Hence a naive Bayes classifier is equivalent to a log-linear model with the two way interactions between each feature $X_k$ and $Y$ and is a log-linear model of conditional independence. In what follows, we parametrize the naive Bayes model using conditional probabilities $P(X_k|Y)$ and the marginal probabilities $P(Y)$. We use a $\square$ to refer to a vector indexed by the indices in place of the box. For instance $n_\square = \{n_1, \ldots, n_I\}$. For an example of a parametrization with $K$ binary features and a binary class $Y$, see Table 2. Thus, let $p_{ij}^k = P(X_k = j | Y = i)$, $p_i = P(Y = i)$ and $n_{ij}^k = \#(Y = i, X_k = j)$. Note that $\sum_{j=1}^{J_k} p_{ij}^k = 1$ for all $i$ and $k$. Similarly, $\sum_{j=1}^{J_k} n_{ij}^k = n_i$ for all $i$ and $k$, where $n_i = \#(Y = i)$. Assume that $[n_{ij}^k]_{j=1}^{J_k} \sim Multinomial(n_i, [p_{ij}^k]_{j=1}^{J_k})$. Similarly, assume that $[n_i] \sim Multinomial(N, [p_i]_{i=1}^{I})$. Let $[p_{ij}^k]_{j=1}^{J_k} \sim Dirichlet([\alpha_{ij}^k]_{j=1}^{J_k})$ and $[p_i]_{i=1}^{I} \sim Dirichlet([\alpha_i]_{i=1}^{I})$ be the priors on the parameters.

Using this notation, the sufficient statistics of the model are $K$ two by two marginal tables of counts $\{n_{ij}^k\}$ for $k = 1, \ldots, K$, see Table 1 for an example. Thus it is sufficient to release these marginals under differential privacy Cormode (2011). We use the Laplace mechanism to release $k$ marginals $\{n_{ij}^k\}$, each marginal can be treated as a histogram query. The global sensitivity, $GS$ of each query is 2 (assuming $N$ is fixed) and hence adding independent Laplace noise with scale parameter $= 2/\epsilon$ to each count in the $k^{th}$ marginal query guarantees $\epsilon$-differential privacy. By composition, releasing all $K$ marginals is $K\epsilon$ differentially private. Hence the released data are $m_{ij}^k = n_{ij}^k + e_{ijk}$, where $e_{ijk} \sim Lap(0, b)$, where $b = \frac{2}{\epsilon}$. As described before, we treat the original data $D = \{n_{ij}^k\}$ as missing and the private counts are $Z = m_{ij}^k$. The parameter vector is $\theta = \{p_{ij}^k, p_i\}$ and we are interested in computing a posterior approximation of the parameters, i.e. $P(\theta|Z)$. This distribution involves an intractable likelihood as we need to sum over all possible tables $\{n_{ij}^k\}$, which is a very large space. Hence we resort to a Variational approximation of the posterior.

### 4.1. Deriving a Variational approximation

To derive a Variational approximation, let us compute the lower bound in equation 1. Recall that $Z = \{m_{ij}^k\}$, $D = \{n_{ij}^k\}$, $\theta = \{p_{ij}^k, p_i\}$ and a $\square$ denotes a vector indexed by the indices in place of the box. Each $m_{ij}^k$ is independently distributed given $n_{ij}^k$ with a Laplace distribution of mean $n_{ij}^k$ and scale parameter $b = \frac{2}{\epsilon}$. Note that $P\left(n_{i\square}^k\right) \sim P\left(n_{i\square}^k|n_i\right) P(n_i)$. For each fixed $i, k$, $P\left(n_{i\square}^k|n_i\right)$ is an independent Multinomial distribution with parameters $p_{i\square}^k$. Finally, $n_\square$ is a multinomial distribution with parameters $p_\square$. Hence, the variational lower bound of the log marginal likelihood $\log L(z)$ is

$$\log L(Z) \geq E_{q(D)q(\theta)}\left[\log \frac{P(Z|D) P(D|\theta) P(\theta)}{q(D)q(\theta)}\right]$$

$$= E\left[\log\left(\prod_{ijk} \frac{P\left(m_{ij}^k|n_{ij}^k\right) P\left(n_{ij}^k|p_{i\square}^k, n_i\right) P\left(p_{ij}^k\right)}{q(n_{i\square}^k|n_i)q(p_{i\square}^k)}\right)\left(\frac{P(n_\square|p_\square)}{q(n_\square)} \frac{P(p_\square)}{q(p_\square)}\right)\right] \stackrel{def}{=} E[\log V].$$

We need to restrict the variational distributions $q(n_{i\square}^k)$ and $q(p_{i\square}^k)$ to a tractable class of distributions so that the expectations can be computed in a closed form. Moreover, we need to choose a distribution on $n_{ij}^k$ that is consistent with the model $P(D|\theta)$, that is the distributions should be such that they imply the same marginal distribution of $Y$ for each $j$ and $k$. To ensure these constraints hold, we define $q(n_{i\square}^k)$ distribution in two steps. Let $q(n_{i\square}^k) = q(n_{i\square}^k|n_\square)q(n_\square)$ where $q(n_{i\square}^k|n_i) = Multinomial(n_i, \theta_{i\square k})$ and $q(n_\square) = Multinomial(N, \theta_\square)$. The distributions $q(p_{i\square}^k)$ and $q(p_\square)$ are unrestricted.

We consider two ways to find a lower bound of the absolute value term due to the Laplace distribution. The first bound is a based on minorizing the absolute value term, (see Hunter and Lange (2004)) by using the concavity of the function $\sqrt{x}$. We call this a *quadratic bound*. Let $\alpha_{ijk}$ be any non-negative number, then

$$-\frac{|m_{ij}^k - n_{ij}^k|}{b} \geq -\frac{1}{2}\left(\frac{(m_{ij}^k - n_{ij}^k)^2}{b\alpha_{ijk}} + \frac{\alpha_{ijk}}{b}\right),$$

with equality holding if and only if $\alpha_{ijk} = |m_{ij}^k - n_{ij}^k|$.

The second bound named *mixture bound* is derived from a mixture representation of the Laplace distribution. Note that the absolute value term is the log kernel of a Laplace random variable with scale parameter $b$. The Laplace random variable can be written as a infinite mixture of Gaussian and Raleigh distributions. Specifically if $P(Z|\beta) \sim N(0, \beta)$ and $P(\beta) \sim Raleigh(b)$, then $P(Z) \sim Laplace(0, b)$, see Proposition 2.2 in Kotz, Kozubowski and Podgorski (2001). This fact combined with Jensen's inequality can be used to bound the absolute value function. It turns out that both the mixture model bound and the quadratic bounds are equivalent up to a re-parametrization. Specifically, if we let $\alpha_{ijk} = \frac{1}{\beta_{ijk}}$, then these two bounds are equivalent to each other. We use the mixture representation based lower bound as it turns out to be computationally stable. After taking expectations and simplifying, the final lower bound is:

$$E[\log V] = \sum_{i=1}^{I}\sum_{ij}^{k} -\frac{3}{2}\mathbb{E}\left[\log \beta_{ijk}\right] - \frac{\mathbb{E}\left[\beta_{ijk}\right]}{2b^2}(N(N-1)\theta_i^2(\theta_{ij}^k)^2 + N\theta_i\theta_{ij}^k + (m_{ij}^k)^2 - 2N\theta_i\theta_{ij}^k m_{ij}^k) - \mathbb{E}\left[\frac{1}{2\beta_{ijk}}\right]$$

$$+\mathbb{E}\left[q(\beta_{ijk})\right] + N\theta_i\theta_{ij}^k\mathbb{E}\left[\log p_{ij}^k\right] - N\theta_i\theta_{ij}^k \log \theta_{ij}^k + \alpha_{ij}^k\mathbb{E}\left[\log p_{ij}^k\right] + \sum_i N\theta_i\mathbb{E}\left[\log p_i\right] - N\theta_i \log \theta_i + \alpha_i\mathbb{E}\left[\log p_i\right]$$

$$-\sum_{ik}\mathbb{E}\left[\log q(p_{i\square}^k)\right] - \mathbb{E}\left[\log q(p_\square)\right]. \quad (2)$$

We need to maximize the lower bound in equation 7 with respect to $\theta_{ij}^k$, $\theta_i$, $q(p_{i\square j})$, $q(\beta_{ijk})$, $q(p_\square)$. Taking the derivatives of equation 7 and setting them equal to 0 gives the following update equations:

$$q(\beta_{ijk}) = InverseGaussian(\lambda = 1, \mu = \frac{b}{\sqrt{k}}) \text{ where } k = \mathbb{E}\left[(m_{ij}^k - n_{ij}^k)^2\right] \quad (3)$$

$$q(p_{i\square}^k) = Dirichlet(\{N\theta_i\theta_{ij}^k + \alpha_{ij}^k + a_i I(j = j_k)\}) \quad (4)$$

$$q(p_\square) = Dirichlet(N\theta_i + \alpha_i + a_i) \quad (5)$$

The derivation is shown in the appendix. Note that we need to take the functional derivative of $q(p_{i\square j}), q(p_\square)$ with the usual constraints that the distribution needs to sum to 1. The optimal solutions to $\theta_{ij}^k$ and $\theta_i$ are obtained by solving the following optimization problems. For each fixed $i$ and $k$,

$$\operatorname*{argmax}_{\theta_{i\square}^k} \sum_j A_j(\theta_{ij}^k)^2 + B_j\theta_{ij}^k + C_j\theta_{ij}^k \log \theta_{ij}^k \quad (6)$$

subject to $\sum_j \theta_{ij}^k = 1$ and $\theta_{ij}^k \geq 0$, where $A_j = \frac{-N(N-1)\theta_i^2\mathbb{E}[\beta_{ijk}]}{2b^2}$, $B_j = \frac{-N\theta_i\mathbb{E}[\beta_{ijk}]}{2b^2} + \frac{Nm_{ij}^k\theta_i\mathbb{E}[\beta_{ijk}]}{b^2} + N\theta_i\mathbb{E}\left[\log p_{ij}^k\right]$, and $C_j = -N\theta_i$. To compute $\theta_\square$, we need to solve

$$\operatorname*{argmax}_{\theta_\square} \sum_i D_i\theta_i^2 + E_i\theta_i + F_i\theta_i \log \theta_i$$

subject to $\sum_i \theta_i = 1$ and $\theta_i \geq 0$. where

$$D_i = -\sum_{jk}\frac{N(N-1)\theta_{ijk}^2\mathbb{E}\left[\beta_{ijk}\right]}{2b^2},$$

$$E_i = \sum_{jk} N\theta_{ij}^k\left(\frac{-\mathbb{E}\left[\beta_{ijk}\right]}{2b^2} + \frac{m_{ij}^k\mathbb{E}\left[\beta_{ijk}\right]}{b^2} + \mathbb{E}\left[\log p_{ij}^k\right] - \log \theta_{ij}^k\right) + N\mathbb{E}\left[\log p_i\right], \text{ and } F_i = -N.$$

We use a first order interior point method to solve these two constrained optimization problems, see Tseng, Bomze and Schachinger (2011). The details of this algorithm are given in the appendix. Note that the interior point method needs careful calibration to ensure that the lower bound always increases, since exact closed form solution is not available. Convergence to the optima of the lower bound is still guaranteed by the theory of MM algorithms where one alternates between Minorizing and Maximizing, see Hunter and Lange (2004). Also note that we did not assume any functional form on the distribution of $\beta_{ijk}$ and the parameters $p_{i\square}^k$ and $p_\square$ and the optimization is performed over all possible distributions. For more details on deriving variational approximations, see Bishop (2006). Some key questions for implementation of the variational approximation remain to be answered, which are addressed next.

*How do we declare convergence?* Determining convergence in this algorithm is not well understood in part because the objective function has many local optimal points. Currently, convergence is declared by monitoring the value of the lower bound to the objective function. We keep track of $E[\log V^t]$ at the $t^{th}$ iteration. We declare convergence when $E[\log V^{t+1}] - E[\log V^t] < tol$ for some pre-specified tolerance value *tol*.

*Choice of starting values.* The choice of starting values is an important tuning parameter in the algorithm. Our experiments show that the number of steps needed for convergence depends on the starting value. A good starting value speeds up convergence. In general, we found that the naive estimates of the conditional class probabilities serve as a good starting point. The naive estimates are defined as those obtained by ignoring the privacy mechanism and using the noisy counts $m_{ij}^k$ as if they were the original counts. In cases where these counts are less than 0 or larger than the total sample size, we simply truncate them to their corresponding upper and lower limits. Finally, we renormalized the counts to make sure that they give a consistent estimate of $p(y)$.

*Selection of priors.* To complete the specification of the algorithm, we need to choose a prior for the parameters $p_{i\square j}$ and $p_\square$. We select the uniform prior on $p_{ij}^k$ and $p_i$.

## 5. Simulation Results

In this section, we evaluate the proposed variational approach on simulated datasets to estimate the approximate posterior distributions of the parameters, i.e. $p_{i\square}^k = \{p(x_k|y = i)\}$ for each feature $k$ and class $i$ and $p_i = p(y = i)$. We use the following method to simulate the data:

1. Generate $p_i = P(Y = i)$ from a Dirichlet distribution with parameters $\alpha_\square$,
2. For each fixed $i$ and $k$, Generate $p_{i\square}^k = P(X_k = j|Y = i)$ from Dirichlet distribution with parameters $\alpha_{i\square k}$,
3. Generate the marginal class counts : $n_i$ from Multinomial$(N, p_i)$,
4. Generate $n_{i\square}^k$ from Multinomial$(n_i, p_{i\square}^k)$.

We compare the mean squared error of three estimators: two private estimators that use the noisy counts $m_{ijk}$ - the naive method that ignores the privacy mechanism (*naive*) and the variational method (*VB*) and a third non-private Bayes estimator (*bayes*) that uses the original counts $n_{ij}^k$ and the uniform prior with $\alpha_\square = 1$. The squared error is calculated between the estimates of $p_{i\square j}$ and $p_\square$ and their true simulated values. The steps used in this study are given below:

Repeat 10 times
1. Generate $n_{i\square}^k$ from Multinomial$(n_i, p_{i\square}^k)$
2. Repeat 5 times
   (a) Add Laplace noise to $n_{ij}^k$ with mean 0 and scale $\frac{2}{\epsilon}$, i.e $m_{ij}^k = n_{ij}^k + e_{ijk}$
   (b) Compute the naive estimates of posterior distribution of $p_{ij}^k$ and $p_i$ using $m_{ij}^k$.
   (c) Compute the variational estimate of posterior distribution using the update equations, until the convergence criteria is met.
   (d) Compute the Bayes estimate of posterior distribution using the true counts $n_{ij}^k$.
   (e) Compute the squared error between the mean parameter estimates and true estimates of $p_{i\square j}$ and $p_\square$
3. End Repeat

In Figure 1 below, we show a box plot of squared error of the estimators of the parameters of the posterior distribution as a function of $\epsilon$ for different sample sizes $N$. Specifically, we vary $\epsilon$ from 0.0001 to 1 and $N \in \{50, 100, 200, 500\}$. The plot clearly shows that the proposed private Variational Bayes estimator beats the naive estimator in terms of the squared error. However, the error of the variational estimator is still higher than the non-private estimator. For very small values of $\epsilon$ and smaller sample sizes, the efficiency (measured by the squared error) gains offered by the variational estimator are much higher when compared to the naive estimator. As $\epsilon$ increases, all the three estimators behave in a similar fashion.

## 6. Future Work

In this paper, we used variational approximations to estimate posterior distributions of the parameters of a naive Bayes model in a private manner. This model is equivalent to a log-linear model with a subset of two-way margins as sufficient statistics. A naive estimator ignores the structure of the contingency table and the noise addition process and uses the noisy counts directly for estimation. However, as we show, using a variational method to impose the structure of the contingency table and modeling the noise addition process in the likelihood leads to reduction in the squared error of parameter estimation. Extension to more general decomposable log-linear models should not pose much difficulty. The challenge would be to choose a parametrization such that the constraints imposed by higher order marginal tables on lower order marginals are satisfied. More work is needed to study the convergence properties of the Variational algorithm proposed in this paper and to understand the effect of starting points on the optimality of the solution. Finally, tighter variational bounds may be used to obtain more accurate approximations. In using the variational approximation, we made an assumption that the distributions $q(n_{ij}^k)$ of $\theta$ and $n_{ij}^k$ are independent. Relaxing this assumption may lead to a more accurate approximation.

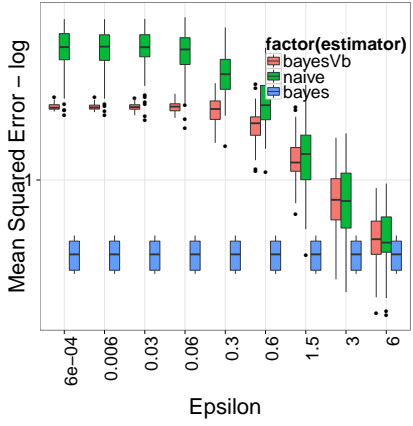
(a) $N = 50$

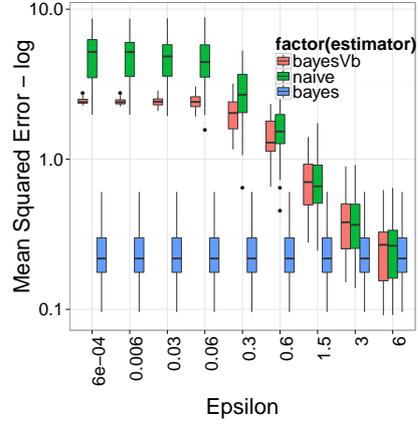
(b) $N = 100$

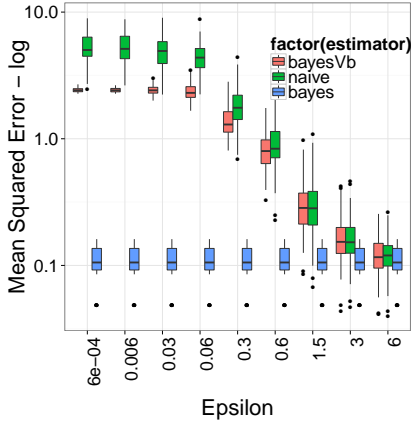
(c) $N = 200$

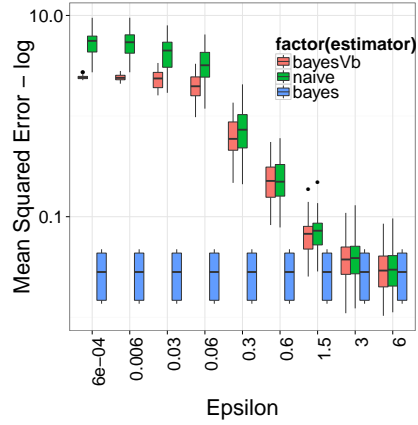
(d) $N = 500$

Fig 1: Comparison of estimators of the posterior distribution using squared error for varying sample size $N$ and $\epsilon$. Here *naive* is the naive estimator based on the noisy counts, *bayesVB* is the variational estimator based on the noisy counts, and *bayes* is the bayes estimator based on the non-private counts.

# 7. Appendix

## 7.1. Derivation of the Variational Lower bound

using the mixture representation of the Laplace distribution and a parametric mean field approximation, the variational lower bound $\log V$ before taking the expectation can be written as follows:

$$\log V = \sum_{ij}^{k} \left( -\frac{3}{2} \log \beta_{ijk} - \frac{\beta_{ijk}}{2b^2}(m_{ij}^k - n_{ij}^k)^2 - \frac{1}{2\beta_{ijk}} - \log q(\beta_{ijk}) \right)$$

$$+ \sum_{ij}^{k} \left( n_{ij}^k \log p_{ij}^k - n_{ij}^k \log \theta_{ij}^k + (\alpha_{ij}^k - 1) \log p_{ij}^k \right)$$

$$+ \sum_{i} n_i \log p_i - n_i \log \theta_i + (\alpha_i - 1) \log p_i$$

$$- \sum_{ik} \log q(p_{i\square}^k)$$

$$- \log q(p_\square).$$

Now, let us take expectation with respect to $q(n_{ij}^k)$, $q(n_i)$ and $q(p_{ij}^k)$, $q(p_i)$. Recall that $n_{i\square}^k | n_i \sim Multinomial(n_i, p_{i\square}^k)$. Hence,

$$\mathbb{E}\left[(m_{ij}^k - n_{ij}^k)^2\right] = (m_{ij}^k)^2 + \mathbb{E}\left[(n_{ij}^k)^2\right] - 2m_{ij}^k \mathbb{E}\left[n_{ij}^k\right]$$
$$= (m_{ij}^k)^2 + Var(n_{ij}^k) + \mathbb{E}\left[n_{ij}^k\right]^2 - 2m_{ij}^k \mathbb{E}\left[\mathbb{E}\left[n_{ij}^k | n_i\right]\right].$$

We can compute $Var(n_{ij}^k)$ by using the conditional variance formula:

$$Var(n_{ij}^k) = \mathbb{E}\left[Var(n_{ij}^k) | n_i\right] + Var(\mathbb{E}\left[n_{ij}^k | n_i\right])$$
$$= \mathbb{E}\left[n_i \theta_{ij}^k (1 - \theta_{ij}^k) | n_i\right] + Var(n_i \theta_{ij}^k)$$
$$= N\theta_i \theta_{ij}^k (1 - \theta_{ij}^k) + (\theta_{ij}^k)^2 N\theta_i (1 - \theta_i)$$
$$= N\theta_i \theta_{ij}^k - N\theta_i^2 (\theta_{ij}^k)^2.$$

Let us also compute $\mathbb{E}\left[n_{ij}^k\right]$ using the conditional expectation formula;

$$\mathbb{E}\left[n_{ij}^k\right] = \mathbb{E}\left[\mathbb{E}\left[n_{ij}^k | n_i\right]\right]$$
$$= N\theta_i \theta_{ij}^k.$$

Thus,

$$\mathbb{E}\left[(m_{ij}^k - n_{ij}^k)^2\right] = (m_{ij}^k)^2 + Var(n_{ij}^k) + \mathbb{E}\left[n_{ij}^k\right]^2 - 2m_{ij}^k \mathbb{E}\left[\mathbb{E}\left[n_{ij}^k | n_i\right]\right]$$
$$= (m_{ij}^k)^2 + N\theta_i \theta_{ij}^k - N\theta_i^2 (\theta_{ij}^k)^2 + N^2 \theta_i^2 (\theta_{ij}^k)^2 - 2m_{ij}^k N\theta_i \theta_{ij}^k$$
$$= N\theta_i^2 (\theta_{ij}^k)^2 (N-1) + N\theta_i \theta_{ij}^k + (m_{ij}^k)^2 - 2m_{ij}^k N\theta_i \theta_{ij}^k.$$

After taking expectations, the final lower bound is:

$$\begin{aligned}
\mathbb{E}\left[\log V\right] =& \sum_{i=1}^{I}\sum_{ij}^{k} -\frac{3}{2}\mathbb{E}\left[\log \beta_{ijk}\right] - \frac{\mathbb{E}\left[\beta_{ijk}\right]}{2b^2}(N(N-1)\theta_i^2(\theta_{ij}^k)^2 + N\theta_i\theta_{ij}^k + (m_{ij}^k)^2 - 2N\theta_i\theta_{ij}^k m_{ij}^k) \\
& - \mathbb{E}\left[\frac{1}{2\beta_{ijk}}\right] + \mathbb{E}\left[q(\beta_{ijk})\right] \\
& + N\theta_i\theta_{ij}^k \mathbb{E}\left[\log p_{ij}^k\right] - N\theta_i\theta_{ij}^k \log \theta_{ij}^k + \alpha_{ij}^k \mathbb{E}\left[\log p_{ij}^k\right] \\
& + \sum_i N\theta_i \mathbb{E}\left[\log p_i\right] - N\theta_i \log \theta_i + \alpha_i \mathbb{E}\left[\log p_i\right] \\
& - \sum_{ik} \mathbb{E}\left[\log q(p_{i\square}^k)\right] \\
& - \mathbb{E}\left[\log q(p_\square)\right].
\end{aligned} \qquad (7)$$

We need to maximize the lower bound in equation 7 with respect to $a_\square$, $\theta_{ij}^k$, $\theta_i$, $q(p_{i\square j})$, $q(\beta_{ijk})$, $q(p_\square)$.

We consider the derivation of the update equations for finding the optimal densities $q(\beta_{ijk})$, $q(p_{ij}^k)$ and $q(p_i)$. We focus on the density $q(\beta_{ijk})$, the derivation of other densities is similar. The optimal density of $\beta_{ijk}$ can be derived by looking at those terms in equation 7 that have $\beta$ in them.

$$\log q(\beta_{ijk}) \propto -\frac{3}{2}\log \beta_{ijk} - \frac{\beta_{ijk}}{2b^2}\mathbb{E}\left[(m_{ij}^k - n_{ij}^k)^2\right] - \frac{1}{2\beta_{ijk}}, \qquad (8)$$

where

$$\mathbb{E}\left[(m_{ij}^k - n_{ij}^k)^2\right] = N(N-1)\theta_i^2(\theta_{ij}^k)^2 + N\theta_i\theta_{ij}^k + (m_{ij}^k)^2 - 2N\theta_i\theta_{ij}^k m_{ij}^k.$$

Note that the inverse Gaussian distribution is given by

$$q(\beta) = \left(\frac{\lambda}{2\pi\beta^3}\right)^{\frac{1}{2}} \exp\left(-\frac{\lambda(\beta-\mu)^2}{2\mu^2\beta}\right). \qquad (9)$$

Now

$$\log q(\beta) \propto -\frac{3}{2}\log \beta - \frac{\lambda\beta}{2\mu^2} - \frac{\lambda}{2\beta}.$$

A comparison between the two log-densities in equation 8 and equation 9 shows that

$$\lambda = 1, \text{ and}$$
$$\frac{\lambda}{\mu^2} = \frac{\mathbb{E}\left[(m_{ij}^k - n_{ij}^k)^2\right]}{b^2}.$$

Thus, the optimal distribution of $\beta_{ijk}$ is an inverse Gaussian distribution with $\lambda = 1$ and the mean parameter,

$$\mu = \frac{b}{\sqrt{\mathbb{E}\left[(m_{ij}^k - n_{ij}^k)^2\right]}}.$$

Note that we did not assume any functional form on the distribution of $\beta_{ijk}$ and the parameters $p^k_{i\square}$ and $p_\square$. Thus, the optimization is performed over all possible distributions. In some of the update equations, such as 3, we need to compute expectations of the form $\mathbb{E}\left[\log p_i\right]$ where $p_i$ is a Dirichlet distribution with parameter vector $\alpha$. This can be computed using the formula: $\mathbb{E}\left[\log p_i\right] = \xi(\alpha_i) - \xi(\sum_i \alpha_i)$ where $\xi$ is the digamma function.

### 7.2. Optimizing $\theta$ parameters

In the derivation of lower bound, a key optimization problem that occurs is of the following type:

$$\operatorname*{argmax}_{\theta_\square} \sum_i A_i \theta_i^2 + B_i \theta_i + C_i \theta_i \log \theta_i,$$

subject to $\sum_i \theta_i = 1$ and $\theta_i \geq 0$, $A_i, C_i \leq 0$. Note that the $A_i$ here is used to denote a coefficient in $\mathbb{R}$ and is different from the set of actions $A$.

Let $\theta$ be the vector of parameters and $f(\theta) = \sum_i A_i \theta_i^2 + B_i \theta_i + C_i \theta_i \log \theta_i$. The constraints on $\theta$ can be expressed succinctly as $\theta \in \triangle$, where $\triangle$ is the simplex. An exact solution to this problem is infeasible. However, in obtaining an approximate solution, care must be taken to ensure that the lower bound is always maximized. This ensures that the overall variational optimization behaves like an MM algorithm and hence converges. Moreover, the speed of convergence depends on how this optimization is performed. There are at least three approaches to solve this problem:

1. Use an MM algorithm to minorize the $\theta \log \theta$ term to make the function quadratic and use a quadratic resource allocation algorithm (Frangioni and Gorgone, 2013) to obtain an exact solution to the minorized optimization;
2. Use an interior point method (Tseng, Bomze and Schachinger, 2011);
3. Use an existing non-linear optimization package.

We use a first order interior point method to solve this problem, see Tseng, Bomze and Schachinger (2011). We briefly explain why the first approach of using an MM algorithm does not work in our setting. In the first approach, we minorize the log function by using the following bound:

$$-\log x \geq -\log y - \frac{1}{y}(x - y),$$

with equality holding if and only if $y = x$. The MM algorithm uses this identity and applies it on terms such as $-n_i \log \theta_i$ and $-n^k_{ij} \log \theta^k_{ij}$. Setting $y$ equal to the previous value of iterate converts this bound into an equality. Hence we can use the following equations to eliminate the log term:

$$-n_i \log \theta_i \geq -n_i \log \theta_i^t - \frac{n_i}{\theta_i^t}(\theta - \theta_i^t)$$

where $\theta^t$ is the value at iteration $t$.

Using this lower bound for the log term transforms the problem into a quadratic optimization problem with the constraints that the solution must lie in a simplex. This problem is well studied in the literature and called a quadratic resource allocation problem, see Frangioni and Gorgone (2013). However, the drawback of this approach is that the solution tends to lie on the boundary, specially in the case of smaller values of $\epsilon$. In such cases, the next iterate becomes undefined due to the appearance of the term $\theta^t$ in the denominator. A black box non-linear optimization routine cannot be used because its solution may not increase the lower bound, and hence convergence may not hold. The interior point method that we explain next avoids both these issues by the use of the $\theta \log \theta$ term, which ensures that the parameters remain away from the boundary and by guaranteeing that the solution produced increases the lower bound.

Let $\theta^{t+1}$ be the value of the parameters at iteration $t+1$. We find a value $\theta^{t+1}$ such that $f(\theta^{t+1}) > f(\theta^t)$, by choosing an appropriate search direction. Let $s^t$ be the step length and $d^t$ be a search direction. Then

$$\theta^{t+1} = \theta^t + s^t d^t, \theta^0 \in \triangle. \tag{10}$$

To define the search direction $d^t$, let $W^t$ be a diagonal matrix with diagonal entires equal to $\theta^t$. Let $e$ be a vector of 1's. Define $d^t = W^t r(\theta^t)$ where

$$r(\theta) = \nabla f(\theta) - \theta^t \nabla f(\theta) e. \tag{11}$$

The step size $s^t$ is restricted such that $0 < s^t < -1/\min_j r(\theta^t)_j$. The restriction on $s^t$ ensures that the optimal point always stays in the simplex. We choose the step size by an Armijo rule (Forsgren, Gill and Wright, 2002), to ensure that the direction in which the solution proceeds is an ascent direction. Specifically, $s^t$ is the largest value in the sequence $s$ where $s \in \{s_0^t(\nu)^k\}_{k=0,1,\ldots}$ such that

$$f(\theta^t + sd^t) \geq f(\theta^t) + \sigma s \nabla f(\theta^t)^T d^t. \tag{12}$$

Here $0 < \nu, \sigma < 1$ are constants and

$$0 < \nu_0^t \begin{cases} \infty & \text{if } d^t \geq 0; \\ -\frac{1}{\min_j d_j^t/\theta_j^t} & \text{otherwise.} \end{cases} \tag{13}$$